# Building Tamil Treebanks[1]

**Kengatharaiyer Sarveswaran**

Department of Computer Science, University of Jaffna, Sri Lanka.

Department of Linguistics, University of Konstanz, Germany.

sarves@univ.jfn.ac.lk

Treebanks are important linguistic resources, which are structured and annotated corpora with rich linguistic annotations. These resources are used in Natural Language Processing (NLP) applications, supporting linguistic analyses, and are essential for training and evaluating various computational models. This paper discusses the creation of Tamil treebanks using three distinct approaches: manual annotation, computational grammars, and machine learning techniques. Manual annotation, though time-consuming and requiring linguistic expertise, ensures high-quality and rich syntactic and semantic information. Computational deep grammars, such as Lexical Functional Grammar (LFG), offer deep linguistic analyses but necessitate significant knowledge of the formalism. Machine learning approaches, utilising off-the-shelf frameworks and tools like Stanza, UDpipe, and UUParser, facilitate the automated annotation of large datasets but depend on the availability of quality annotated data, cross-linguistic training resources, and computational power. The paper discusses the challenges encountered in building Tamil treebanks, including issues with Internet data, the need for comprehensive linguistic analysis, and the difficulty of finding skilled annotators. Despite these challenges, the development of Tamil treebanks is essential for advancing linguistic research and improving NLP tools for Tamil.

## 1.0 Introduction

Treebanks are important because they provide structured, annotated corpora that serve as crucial resources for training and evaluating Natural Language Processing (NLP) models. They offer detailed syntactic and sometimes semantic information, which helps in understanding the grammatical structure of languages.

These treebanks are then used to build tools called parsers, which are employed to parse sentences and obtain their syntactic analyses. Parsing is considered a core task in NLP and is crucial for enabling computational models to understand the syntax of a language.

Treebanks support the development of parsers, which are essential for applications like machine translation, sentiment analysis, and information extraction. For instance, the Penn Treebank provides detailed part-of-speech tagging, hierarchical structures capturing syntactic relationships, and a clear demarcation of phrases (e.g., noun phrases, verb phrases, prepositional phrases).

A sentence such as "The teacher explained the complex topic to the students" can be annotated in a Penn Treebank-inspired way as follows:

---

[1]Invited talk at ICTCIT 2024, held at the University of Texas at Dallas from June 14-16, 2024.
Sarveswaran, K. (2024). *Building Tamil Treebanks*. In *Proceedings of the International Conference on Tamil Computing and Information Technology (ICTCIT 2024)/23rd Tamil Internet Conference* (pp. 22-32). INFITT. ISSN: 2313-4887.

```
(S
  (NP (DT The) (NN teacher))
  (VP (VBD explained)
      (NP (DT the) (JJ complex) (NN topic))
      (PP (TO to)
      (NP (DT the) (NNS students)))))
```

In this structure, NN, DT, JJ, etc., are part-of-speech tags. The bracket structure shows the hierarchical structure of the sentence. NP, VP, and PP are phrases. Although the Penn Treebank provides a framework for English syntactic annotation, it can be adapted to Tamil or other languages with specific adjustments to account for linguistic differences. For instance, the phrases sometimes need to be reordered to fit the structure. The Tamil translation of the sentence given above is shown in (1), and the syntactic tree for that would be (2).

(1) ஆசிரியர் சிக்கலான தலைப்பை மாணவர்களுக்கு விளக்கினார்
*āciriyar cikkalāṉa talaippai māṇavarkaḷukku viḷakkiṉār*
teacher complex.Adj topic.Acc student.Pl.Dat explain.Past.3SgEpi
"The teacher explained the complex topic to the students"

(2)
```
(S
  (NP (NN ஆசிரியர்))
  (VP (VB விளக்கினார்)
      (NP (JJ சிக்கலான) (NN தலைப்பை))
      (PP (TO மாணவர்களுக்கு))))
```

These annotated data for English and Tamil are useful for performing phrase structure alignment, and therefore, useful for developing machine translation applications, for instance.

Treebanks also facilitate linguistic research by providing data that can be used to test linguistic theories and hypotheses. Additionally, treebanks enable cross-linguistic studies and the comparison of syntactic phenomena across different languages. For instance, Futrell (2015) shows that Tamil has the highest degree of free subject and object order using a cross-linguistic analysis from treebanks available in the Universal Dependencies repository.

Although large language models (LLMs) are mostly trained using raw text, treebanks are crucial for evaluating whether LLMs capture syntactic nuances of languages. For instance, Tenney et al. (2019) evaluate the Bidirectional Encoder Representations from Transformers (BERT) model using a treebank to check its linguistic capabilities.

In summary, treebanks are essential resources for training and evaluating computational models and conducting linguistic studies.

In this paper, I discuss how we built Tamil treebanks using various approaches and some of the challenges encountered. The paper consists of three main sections: Building Treebanks, Discussion and Conclusion.

## 2.0 Treebank annotation formats

Several annotation schemes are used in different treebanks, including the Penn Treebank annotation (Marcus, Marcinkiewicz, & Santorini, 1993), the Prague Dependency Treebank (PDT) annotation (Hajič et al., 2001), the Paninian Dependency framework (Bharati & Sangal, 1993; Begum et al., 2008), and the Universal Dependencies annotation (Nivre et al., 2016). Each of these schemes captures various levels of syntactic and sometimes semantic information and is backed by various linguistic theories.

The primary classification of these schemes can be divided into phrase structure, as followed in resources like the Penn Treebank, and dependency schemes, as used in Universal Dependencies. Analysing the pros and cons of these schemes is beyond the scope of this article.

Among the available formalisms, the dependency grammar formalism is particularly useful for languages such as Tamil, which are morphologically rich and have relatively variable and less rigid word order (Bharati et al., 2009). Since sentences can be constructed with various word orders, phrase structure rules can easily break down. To accurately capture even a simple sentence, multiple phrase structure rules are often required.

Let's take the example sentence in Tamil as shown in (03).

(03) [வெளியுறவுத்துறை அமைச்சர்] [மரியாதை நிமித்தமாக] [அன்வரை] [சந்தித்தார்]
*veḷiyuṟavuttuṟai amaiccar mariyātai nimittamāka aṉvarai cantittār*
foreign-depart.Nom minister.Nom courtesy out-of.Adv anwar.Acc meet.Past.3SgEpi
Foreign Minister paid a courtesy call on anwar

The sentence in (03) is in the form of $NP_{subj}$ ADVP $NP_{obj}$ V, and this phrase order can be changed in Tamil, for instance, to $NP_{obj}$ $NP_{subj}$ ADVP V or ADVP $NP_{subj}$ $NP_{obj}$ V. To effectively capture this simple sentence, you would need to write at least three different rules in the phrase structure formalism.

On the other hand, a single dependency structure can capture this free-word order nature, as shown in the figure below. Even if the word-order changes, the same dependency rules apply.

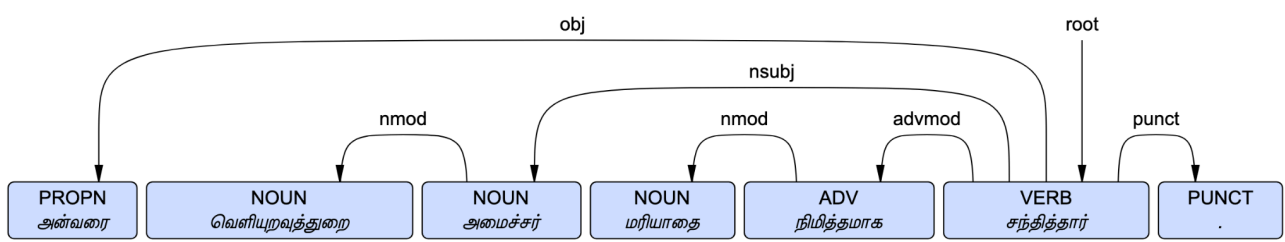

Figure 01: The Universal Dependencies based dependency structure for (03)

There are other linguistically rich and deep grammar formalisms also used to create dependency treebanks. For instance, Lexical-Functional Grammar (LFG) (Kaplan & Bresnan, 1981) provides both a constituency (c-structure) and a dependency (functional structure or f-structure) representation. Figure 02 shows an analysis of a simple construction using the LFG formalism. As illustrated in Figure 02, the constituency structure (c-structure) is represented in the form of a tree, while the dependency structure is shown as an attribute-value matrix in the functional structure (f-structure). Even if the word order changes, the f-structure remains the same, whereas the c-structure will change.

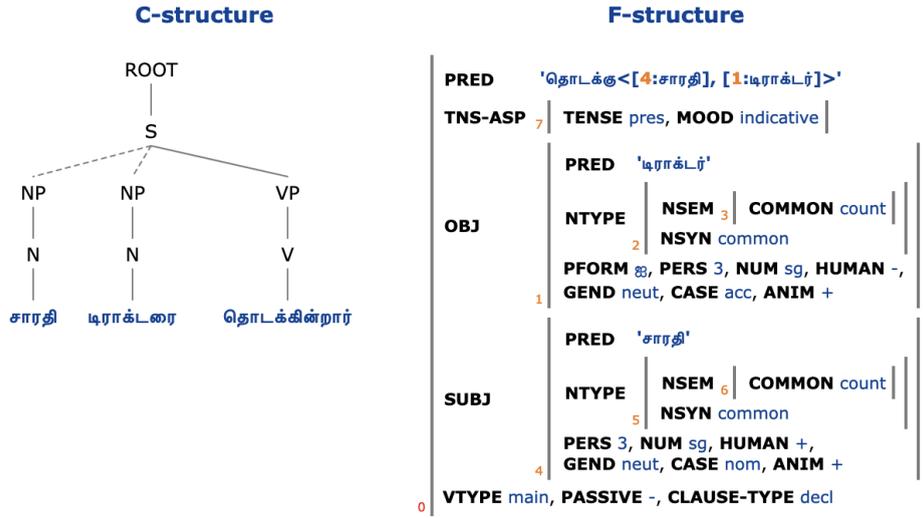

சாரதி டிராக்டரைத் தொடக்கின்றார்
*cārati ṭirākṭarait toṭakkiṉṟār*
driver tractor.Acc start.Pres.3SgEpi
The driver starts the tractor

Figure 02: An example for a Lexical Functional Grammar analysis

## 2.3 Building treebanks

This section outlines the creation of Tamil treebanks using three distinct approaches: manual annotation using the Universal Dependencies, computational grammars, and machine learning techniques. There are several annotation schemes available, as mentioned earlier. In this section, I will discuss the schemes I used to annotate treebanks, as examples.

### 2.3.1 Building treebanks using manual annotation

Treebanks can be created manually by trained annotators and linguists. For instance, the Modern Written Tamil Treebank and the Aalamaram Tamil Treebank were created manually. This is a time-consuming and tedious process, requiring annotators to have extensive training and an in-depth understanding of Tamil linguistics.

The typical manual annotation process involves selecting and cleaning the data. Then, annotators are trained to annotate the data using an annotation guideline. For languages like Tamil, such guidelines may not exist initially and need to be bootstrapped. Typically, an initial version of the guidelines is taken from another language and then iteratively adapted and refined for Tamil during the annotation process.

Currently, the Universal Dependencies (UD) framework is widely used to build treebanks for various languages. In the latest version of Universal Dependencies, 283 treebanks have been created using 161 languages worldwide. Some of these treebanks are created based on specific themes. For instance, there is a Vedic treebank created by an institute in Switzerland.

At least four efforts have been made to create a Universal Dependencies treebank for Tamil. One such treebank was initially created using another dependency scheme called the Prague Dependency Treebank (Ramasamy, 2011), and then it was converted automatically using a script. Therefore, there are some flaws in the treebank. The Modern Written Tamil Treebank (MWTT) by Krishnamurthy and Sarveswaran (2021) was created using examples extracted from grammar books. This MWTT consists of 600 sentences. Recently, two other large-scale Universal Dependencies treebanks, each with roughly 100,000 tokens, have been created by two groups of researchers (Abirami et al., 2024).

Table 01 shows the basic information captured in Universal Dependencies. However, all this information can be extended to capture language-specific features. The initial set of features was proposed by researchers considering cross-lingual and multilingual processing. For instance, Abirami et al. (2024) extended this specification to capture Named Entities in the Tamil treebank. The authors used the Misc field to include NER annotation. This field was previously used to include the transliteration of forms and lemmas of the respective sentence.

Table 01: An example of the Universal Dependency annotation (CoNLL-U format)

| ID | Form | Lemma | POS | XPOS | Morph-features | Rel | Deprel | Misc |
|---|---|---|---|---|---|---|---|---|
| 1 | பையன் | பையன் | NOUN | _ | Case=Nom\|Gender=Masc\|Number=Sing\|Person=3 | 4 | nsubj | — |
| 2 | சாவியால் | சாவி | NOUN | _ | Case=Ins\|Gender=Neut\|Number=Sing\|Person=3 | 4 | obl:inst | — |
| 3 | கதவைத் | கதவு | NOUN | _ | Case=Acc\|Gender=Neut\|Number=Sing\|Person=3 | 4 | obj | — |
| 4 | திறந்தான் | திற | VERB | _ | Gender=Masc\|Mood=Ind\|Number=Sing\|Person=3\|Polarity=Pos\|Tense=Past\|VerbForm=Fin\|Voice=Act | 0 | root | — |
| 5 | . | . | PUNCT | _ | PunctType=Peri | 4 | punct | — |

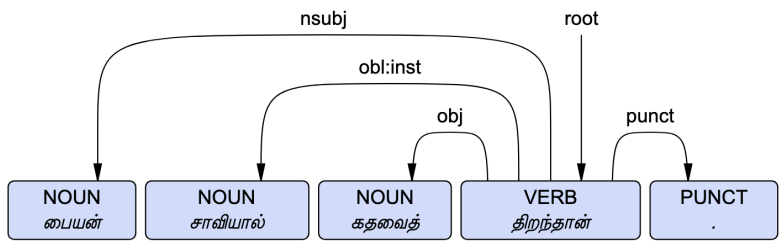

Figure 03: A dependency graph for the annotation given in Table 01.

### 2.3.2 Building treebanks using Computational Grammars

Grammar-based treebank development is not very popular because it requires significant linguistic knowledge, familiarity with grammar formalisms, and the ability to model these formalisms using computers to generate treebanks. These grammar-based annotations are primarily used for linguistic analyses and comparative studies. There are a few modern linguistic formalisms, such as Head-Driven Phrase Structure Grammar (HPSG) and Lexical Functional Grammar (LFG), that are actively used to develop grammars for various languages. These two formalisms are also called deep grammar formalisms because they can model deep syntactic structures that are invariant among languages and provide ways to handle various syntactic transformations using syntactic rules, such as passive and dative shifts.

An effort has been made to build a Tamil Lexical Functional Grammar, which is still a work in progress. Currently, the grammar has been implemented to parse very simple sentences taken from elementary Tamil books. Additionally, sentences were obtained from the Parallel Grammar project (ParGram)(Butt et al, 2002), which aims to develop parallel LFG grammars for several languages worldwide to support cross-lingual analyses.

Lexical Functional Grammar is a useful formalism and is now being widely used for various levels of linguistic analysis beyond morphosyntax, including prosody and semantics. However, there is still a long way to go in this respect for Tamil grammar. The environment used to write LFG grammars is called the Xerox Linguistic Environment (XLE). In this environment, phrase structure rules, lexical rules, and lexicon entries are used to build the complete grammar. It is important to note that the XLE environment also supports the integration of morphological analysers developed using Finite-State Transducers (FST), which increases the robustness of the grammar.

Once the grammar is in place, it can be converted into a parse bank (Sulger et al, 2013), which consists of all the annotations along with the sentences. Platforms like the Infrastructure for the Exploration of Syntax and Semantics (INESS) host such parse banks and treebanks, providing access to parallel analyses (Rosén, 2012). There have also been attempts to convert parse banks generated using Lexical Functional Grammar into other treebank formats, such as the Universal Dependencies.

### 2.3.3 Building Treebanks Using Machine Learning Approaches

Treebanks can also be built using machine learning or deep learning approaches. Several off-the-shelf tools are available for building treebanks using Universal Dependencies parsing, including Stanza, UDpipe, and UUParser. These tools can annotate given sentences using the Universal Dependencies framework. However, to train these parsers, annotated Tamil data is necessary. One approach, called multilingual parsing, helps train a parser for a language using data from similar languages. The effectiveness of this method depends on the accuracy of the data from the other languages.

We built a deep learning-based parser called *Thamizhi*UDp (Sarveswaran & Dias, 2020) using deep learning and a widely used low-resource language processing technology called multilingual processing. Instead of using a machine learning or deep learning approach to annotate the treebank end-to-end, we employed various tools to create the annotations through a multi-stage approach.

First, existing POS-tagged corpora were collected, and a POS tagger was trained using available POS-tagged data to perform part-of-speech tagging with Stanza (Peng. et al, 2020). Then, based on the POS information, ThamizhiMorph (Sarveswaran et al., 2021), a finite-state morphological analyzer, was used to include UD morphological features in the data. Subsequently, a UUParser-based multilingual parser (Smith, 2018) trained using Hindi, Arabic, Telugu, and Turkish languages was used to annotate the dependency information. We also manually annotated a dataset to validate the parser for syntactic coverage and the parsing accuracy.

Data quality and the amount plays a crucial role in multilingual parsing. For example, although Telugu is in the same language family as Tamil, the parser gave better results when trained with Hindi data, as the Hindi treebank contains a substantial amount of manually annotated data. Some of these experiments were reported in Sarveswaran & Dias (2020).

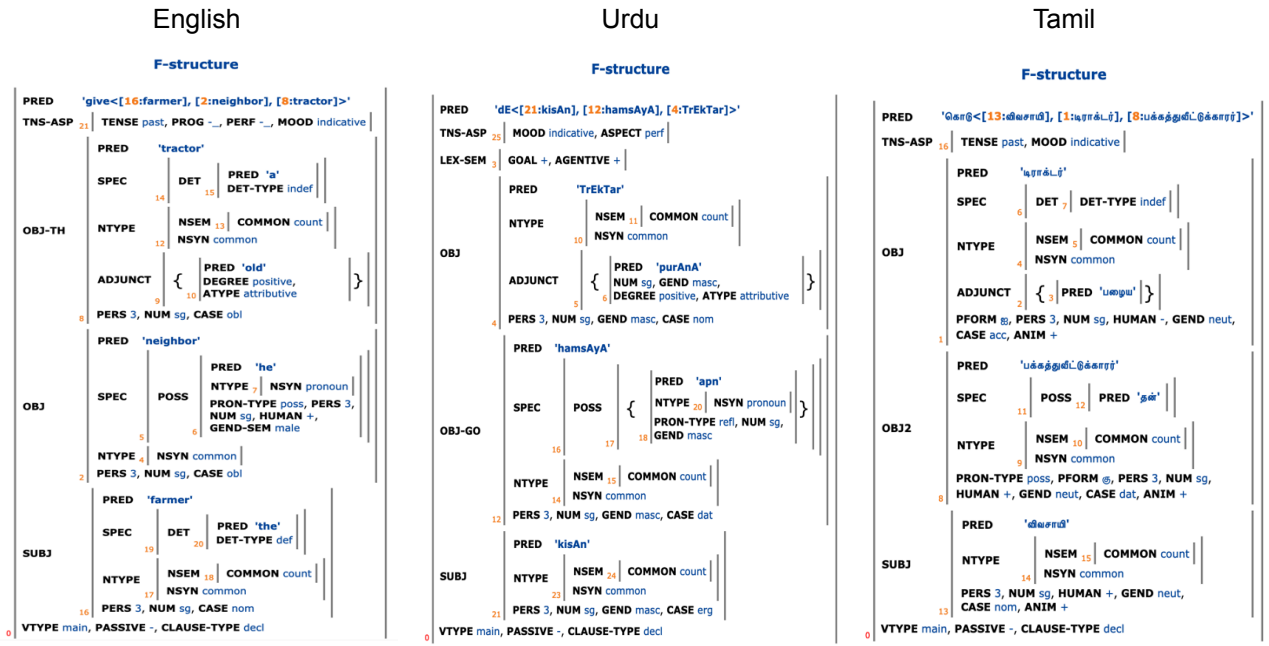

Figure 04: Parallel analyses for "My neighbor was given an old tractor by the farmer" in English, Urdu and Tamil using the Lexical Functional Grammar formalism

Several off-the-shelf Universal Dependency parsers are also available online for Tamil. However, since these are trained using the data available online, the quality of the analysis is often very poor.

## 4.0 Discussion

We faced numerous challenges while building Tamil treebanks. This section highlights two such challenges related to data and the annotation process.

### 4.1 The Nature of Internet Data

The sentences taken from grammar books were well-structured and clean. However, the data found on the Internet presented several issues, including the following:

- Code-mixed Data: Many sentences include a mix of Tamil and other languages, complicating the parsing and annotation processes. In addition to the language code-mix, there was also junk data found in the dataset, such as HTML tags.

- Spelling and Grammar Mistakes: Internet data often contains many spelling and grammatical errors that can complicate the annotation process. Annotators always struggle to get the context in such cases and decide whether to annotate incorrect sentences or not. However, sometimes it is also important to annotate incorrect sentences to train the model effectively.

- Fragments: Incomplete sentences or sentence fragments are common, making it difficult to provide accurate syntactic and/or semantic annotations.

- Dialects: Tamil has various dialects, some of which are not easily understandable or standardised, posing additional challenges for consistent annotation. It also depends on the treebank design whether to annotate dialects or not.

While these problems are common for any application development using Internet data, they are particularly challenging for treebank creation, where we seek linguistic insights. For instance, syntactic ambiguities can lead to multiple parse trees. Handling code-mixed data is also challenging.

## 4.2 Linguistic Analysis

This is the most significant challenge we face when building treebanks. Annotators need an in-depth linguistic understanding to accurately annotate sentences. However, since there is no modern comprehensive grammar available for Tamil, identifying analyses can be difficult. For instance, light verb constructions are challenging because identifying and annotating their parts in Tamil can be complex for annotators. Additionally, mixed categories present difficulties; certain constructions in Tamil, such as *vinaiyaalannaiyum peyar*, exhibit both nominal and verbal features, complicating their categorisation and annotation (Butt et al., 2020). Handling and annotating these requires deep linguistic understanding.

Very long sentences found in formal writing are also problematic. For instance, we encountered sentences with more than 40 tokens. When the number of tokens increases, marking dependencies becomes very challenging for humans, even though most of these are central embeddings or modifiers.

Tamil words often contain packed linguistic information. To annotate them, we need to break them into individual pieces that can be marked for syntactic information. For example, we tokenize clitics from the forms to mark their syntactic roles. For instance, Figure 05 shows how *-um* is tokenised to mark concessiveness and the conjunction.

Furthermore, people tend to stack more and more tokens to form compounds, complicating language processing. In such cases, we need to break them into multiple tokens to capture their syntactic information. However, breaking such tokens is not straightforward due to the nature of Unicode encoding and the abugida writing system. Additionally, deep linguistic knowledge is required to break them accurately.

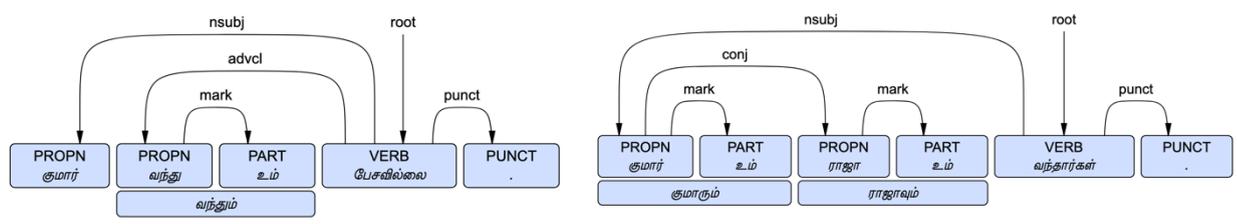

குமார் வந்தும்    பேசவில்லை  
kumār vantālum pēcavillai  
Kumar come.Past.Consessive speak-no.  
Although Kumar came, he did not speak.

குமாரும்    ராஜாவும் வந்தார்கள்  
kumārum rājāvum    vantārkaḷ  
kumar.and raja.ana came  
Kumara and Raja Came.

Figure 05: Concessive and Conjunction functions of 'um'

Identifying good annotators passionate about linguistic annotation is very challenging. While we can find native Tamil speakers, not many have an understanding of linguistic phenomena or it was hard to train them. Moreover, many linguists today prefer other branches of linguistics, and few are interested in studying language structure.

Despite these challenges, the development of Tamil treebanks is crucial for advancing linguistic research and improving NLP tools for Tamil. The efforts to overcome these challenges contribute to more robust linguistic resources.

## 5.0 Conclusion

In this paper, I have briefly outlined three approaches for building treebanks: manual annotation, grammar-based parsing, and machine learning approaches. Each of these methods captures different levels of information. However, the amount of information they can capture can be expanded using language-specific features. A significant advantage of formalisms and annotation schemes discussed is their expandability. Among these, Lexical Functional Grammar (LFG)-based parsed treebanks are highly deterministic and built on a solid linguistic foundation. Therefore, they are particularly useful for linguistic analyses.

Furthermore, these treebanks are crucial in the current era of Large Language Models (LLMs) as they can be used to fine-tune and evaluate these models, making the models more effective for a wide variety of tasks. Although there have been some efforts reported in creating treebanks, there is still a long way to go. More studies related to Tamil linguistics are needed, especially to capture contemporary Tamil and dialectal variations.

## Acknowledgement

Some of these reported works were carried out in collaboration with many scholars around the world, including Miriam Butt, Gihan Dias, Parameswari Krishnamurthy, Keerthana Balasubramani, A. M. Abirami, Wei Qi Leong, Hamsawardhini Rengarajan, D. Anitha, R. Suganya, and many others.

## References:

Abirami, A. M., Leong, W. Q., Rengarajan, H., Anitha, D., Suganya, R., Singh, H., … Shah, R. R. (2024, May). Aalamaram: A Large-Scale Linguistically Annotated Treebank for the Tamil Language. In G. N. Jha, S. L., K. Bali, & A. K. Ojha (Eds.), *Proceedings of the 7th Workshop on Indian Language Data: Resources and Evaluation* (pp. 73–83).

Begum, R., Husain, S., Dhwaj, A., Sharma, D. M., Bai, L., & Sangal, R. (2008). Dependency annotation scheme for Indian languages. In *Proceedings of the Third International Joint Conference on Natural Language Processing: Volume-II.*

Bharati, A., & Sangal, R. (1993, June). Parsing free word order languages in the Paninian framework. In *31st Annual Meeting of the Association for Computational Linguistics* (pp. 105-111)

Bharati, A., Gupta, M., Yadav, V., Gali, K., & Sharma, D. M. (2009, August). Simple parser for Indian languages in a dependency framework. In *Proceedings of the Third Linguistic Annotation Workshop (LAW III)* (pp. 162-165).

Butt, M., Rajamathangi, S., & Sarveswaran, K. (2020). Mixed Categories in Tamil via Complex Categories. In M. Butt & I. Toivonen (Eds.), *Proceedings of the LFG'20 Conference, On-Line* (pp. 68–88).

Butt, M., Dyvik, H., King, T., Masuichi, H., & Rohrer, C. (2002). The Parallel Grammar Project. In *COLING-02: Grammar Engineering and Evaluation*.

Futrell, R., Mahowald, K., & Gibson, E. (2015, August). Quantifying word order freedom in dependency corpora. In *Proceedings of the third international conference on dependency linguistics (Depling 2015)* (pp. 91-100).

Kaplan, R. M., & Bresnan, J. (1981). *Lexical-functional grammar: A formal system for grammatical representation*. Massachusetts Institute Of Technology, Center For Cognitive Science.

Krishnamurthy, P., & Sarveswaran, K. (2021, December). Towards building a modern written tamil treebank. In *Proceedings of the 20th International Workshop on Treebanks and Linguistic Theories (TLT, SyntaxFest 2021)* (pp. 61-68).

Peng Qi, Yuhao Zhang, Yuhui Zhang, Jason Bolton, and Christopher D Manning. 2020a. Stanza: A python natural language processing toolkit for many human languages. In Proceedings of the *58th Annual Meeting of the Association for Computational Linguistics: System Demonstrations*, pages 101–108.

Ramasamy, L., & Žabokrtský, Z. (2012, May). Prague dependency style treebank for Tamil. In *Proceedings of the Eighth International Conference on Language Resources and Evaluation (LREC'12)* (pp. 1888-1894).

Sarveswaran, K., & Dias, G. (2020, December). *Thamizhi*UDp: A Dependency Parser for Tamil. In P. Bhattacharyya, D. M. Sharma, & R. Sangal (Eds.), *Proceedings of the 17th International Conference on Natural Language Processing (ICON)* (pp. 200–207).

Sarveswaran, K., Dias, G., & Butt, M. (2021). *Thamizhi*Morph: A morphological parser for the Tamil language. *Machine Translation*, *35*(1), 37-70.

Smith, A., Bohnet, B., Nivre, J., de Lhoneux, M., Stymne, S., & Shao, Y. (2018). 82 treebanks, 34 models: Universal dependency parsing with cross-treebank models. Retrieved from https://research.google/pubs/82-treebanks-34-models-universal-dependency-parsing-with-cross-treebank-models/

Sulger, S., Butt, M., King, T., Meurer, P., Laczko, T., Rákosi, G., Dione, C., Dyvik, H., Rosén, V., De Smedt, K., Patejuk, A., Çetinoğlu, ., Arka, I., & Mistica, M. (2013). ParGramBank: The ParGram Parallel Treebank. In *Proceedings of the 51st Annual Meeting of the Association for Computational Linguistics (Volume 1: Long Papers)* (pp. 550–560). Association for Computational Linguistics.

Tenney, I., Das, D., & Pavlick, E. (2019). BERT rediscovers the classical NLP pipeline. *arXiv preprint arXiv:1905.05950*.

Victoria Rosén, Koenraad De Smedt, Paul Meurer, and Helge Dyvik. An open infrastructure for advanced treebanking. In Jan Hajič, Koenraad De Smedt, Marko Tadić, and António Branco (eds.) *META-RESEARCH Workshop on Advanced Treebanking at LREC2012,* pages 22–29, Istanbul, Turkey, May 2012.